\documentclass[journal]{IEEEtran}
\ifCLASSOPTIONcompsoc
  \usepackage[nocompress]{cite}
\else
  \usepackage[nocompress,sort]{cite}
\fi
\usepackage{times}
\usepackage{epsfig}
\usepackage{graphicx}
\usepackage{tabulary}
\usepackage{tabularx}
\newcommand{\specialcellL}[2][c]{%
  \begin{tabular}[#1]{@{}l@{}}#2\end{tabular}}

\usepackage{amsmath}
\usepackage{amssymb}
\let\labelindent\relax
\usepackage[shortlabels]{enumitem}
\usepackage[normalem]{ulem}
\usepackage{multirow}
\usepackage{threeparttable}
\usepackage[tight,normalsize,sf,SF]{subfigure}
\usepackage[pagebackref=true,breaklinks=true,colorlinks,bookmarks=false]{hyperref}

\newcommand{\modelnamelong}[0]{\textit{Region Convolutional 3D Network (R-C3D) }}
\newcommand{\modelname}[0]{R-C3D }

\begin{document}
%
\title{Two-Stream Region Convolutional 3D Network for Temporal Activity Detection}
%
%
%
%

\author{Huijuan~Xu\IEEEauthorrefmark{1},~
        Abir~Das~
        and~Kate~Saenko~
\IEEEcompsocitemizethanks{\IEEEcompsocthanksitem \IEEEauthorrefmark{1} Corresponding author.
\IEEEcompsocthanksitem H. Xu is at UC Berkeley, A. Das is at IIT Kharagpur and K. Saenko is at Boston University. H. Xu and A. Das were at Boston University at the time of this work.
\IEEEcompsocthanksitem E-mails: huijuan@eecs.berkeley.edu, abir@cse.iitkgp.ac.in, saenko@bu.edu}
\thanks{Manuscript received January 15, 2018; revised April 12, 2019.}}

\markboth{Journal of \LaTeX\ Class Files,~Vol.~6, No.~1, April~2019}%
{Shell \MakeLowercase{\textit{et al.}}: Bare Demo of IEEEtran.cls for Computer Society Journals}

\IEEEcompsoctitleabstractindextext{%
\begin{abstract}
We address the problem of temporal activity detection in continuous, untrimmed video streams.
This is a difficult task that requires extracting meaningful spatio-temporal features to capture activities, accurately localizing the start and end times of each activity.
We introduce a new model, \textit{Region Convolutional 3D Network (R-C3D)}, which encodes the video streams using a three-dimensional fully convolutional network, then generates candidate temporal regions containing activities and finally classifies selected regions into specific activities.
Computation is saved due to the sharing of convolutional features between the proposal and the classification pipelines.
We further improve the detection performance by efficiently integrating an optical flow based motion stream with the original RGB stream.
The two-stream network is jointly optimized by fusing \textcolor{black}{the flow and RGB feature maps at different levels.}
Additionally, the training stage incorporates an online hard example mining strategy to address the extreme foreground-background imbalance typically observed in any detection pipeline.
Instead of heuristically sampling the candidate segments for the final activity classification stage, we rank them according to their performance and only select the worst performers to update the model.
\textcolor{black}{This improves the model without heavy hyper-parameter tuning.}
\textcolor{black}{Extensive experiments on three benchmark datasets are carried out to show superior performance over existing temporal activity detection methods.}
Our model achieves state-of-the-art results on the THUMOS'14 and Charades datasets.
We further demonstrate that our model is a general temporal activity detection framework that does not rely on assumptions about particular dataset properties by evaluating our approach on the ActivityNet dataset.
\end{abstract}

}

\maketitle

\IEEEdisplaynotcompsoctitleabstractindextext

%
\IEEEpeerreviewmaketitle

\section{Introduction}

\IEEEPARstart{V}{ideo} scene understanding is an important computer vision problem with many practical applications, including smart
surveillance, monitoring of patients or elderly, online video retrieval \textit{etc}.
Over the past few years, it has quickly evolved from classifying a short, trimmed video to detecting multiple activities in long, untrimmed videos.
This is a more challenging problem compared to trimmed video classification as it requires not only recognizing, but also precisely localizing activities in time.
Most of the existing works rely on a large set of features and separate classifiers exhaustively applied to a set of video segments extracted from the input video using sliding windows~\cite{karaman2014fast, oneata2014lear, shou2016temporal, wang2014action}.
These approaches suffer from one or more of the following major drawbacks: they do not learn deep representations in a jointly optimized fashion, but rather use hand-crafted features~\cite{wang2013action, wang2014video}, or employ deep features like VGG~\cite{simonyan2014very}, ResNet~\cite{he2016deep}, C3D~\cite{tran2015learning} \textit{etc.}, learned separately on image/video classification tasks.
Such off-the-shelf representations may not be optimal for localizing activities in diverse video domains, resulting in inferior performance.
Furthermore, current methods' dependence on external proposal generation or exhaustive sliding windows leads to poor computational efficiency.
Finally, the sliding-window models cannot easily predict flexible activity boundaries due to the fixed temporal granularity of the sliding windows.

\begin{figure}[t]
\centering
\includegraphics[width=1.0\linewidth]{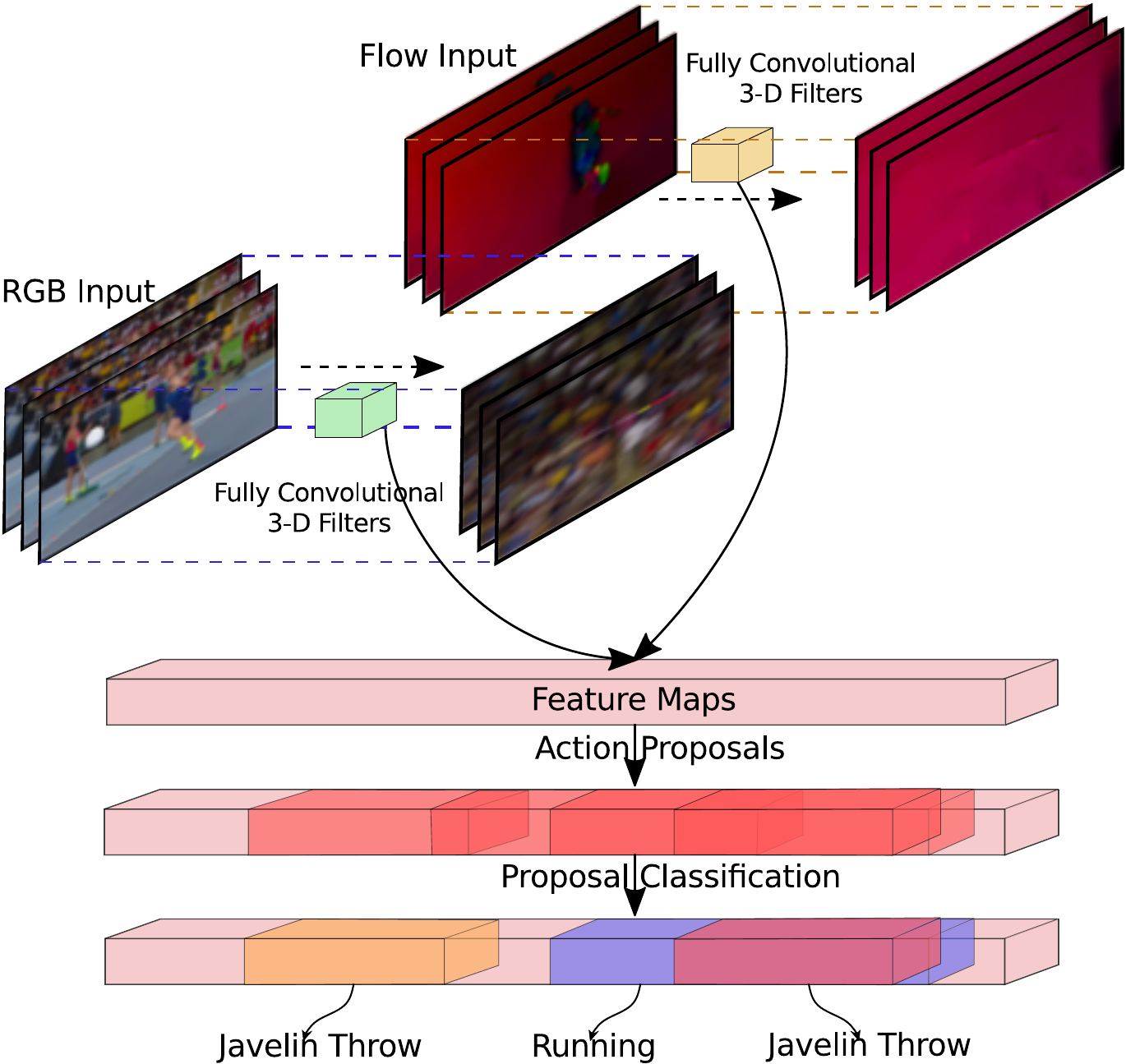}
\caption{We propose a fast two-stream \modelnamelong for temporal activity detection in continuous videos. It encodes both the RGB frames and the optical flow maps in two separate streams with fully-convolutional 3D filters, proposes activity segments, then classifies and refines them based on pooled features within the segment boundaries. Our model improves both speed and accuracy compared to existing methods.}
\label{fig:schematic}
\end{figure}

In this paper, we propose an activity detection model that addresses the above issues. Our \modelnamelong is jointly trainable and learns task-dependent convolutional features by jointly optimizing proposal generation and activity classification.
Inspired by the Faster R-CNN~\cite{ren2015faster} object detection approach, we compute fully-convolutional 3D ConvNet features and propose temporal regions likely to contain activities, then pool features within the proposals to predict activity classes (Figure~\ref{fig:schematic}).
The proposal generation stage filters out many background segments with superior computational efficiency compared to sliding window models.
Furthermore, proposals are predicted with respect to predefined anchor segments and can be of variable length, allowing detection of activities with flexible boundaries.

Convolutional Neural Network (CNN) features learned end-to-end have been  successfully used for activity recognition~\cite{karpathy2014large, simonyan2014two}, particularly in 3D ConvNets (C3D~\cite{tran2015learning}), which learns to capture spatio-temporal features.
However, unlike the traditional usage of 3D ConvNets~\cite{tran2015learning} where the input is short 16-frame video chunks, our method applies full convolution along the temporal dimension to encode as many frames as the GPU memory allows. Thus, rich spatio-temporal features are automatically learned from longer videos.
These feature maps are shared between the activity proposal  and classification subnets to save computation time and jointly optimize features for both tasks.

Alternative activity detection approaches~\cite{escorcia2016daps, montes2016temporal, Singh2016a, yeung2016end} use a recurrent neural network (RNN) to encode sequence of frames or  video chunk features (VGG~\cite{simonyan2014very}, C3D~\cite{tran2015learning}) and predict the activity label at each time step. However, these RNN based methods can only model temporal features at a fixed granularity (e.g. per-frame CNN features or 16-frame C3D features). 
In order to use the same classification network to classify variable length proposals into specific activities, we extend 2D region of interest (RoI) pooling to 3D which extracts a fixed-length feature representation for these proposals. Thus, our model can utilize video features at any temporal granularity.
Furthermore, some RNN-based detectors rely on direct regression to predict the temporal boundaries.
As shown in object detection~\cite{girshick2014rich, Szegedy2013} and semantic segmentation~\cite{Carreira2012}, object boundaries obtained using a regression-only framework are inferior compared to `proposal based detection'.

Motion information plays a pivotal role in video scene understanding.
While deep CNN features have worked remarkably well in combating a multitude of spatial variations due to changes in lighting, pose, scale \textit{etc}., the performance of 3-D CNNs alone for large-scale video understanding is, still, limited~\cite{Diba2016, Carreira2017, simonyan2014two, Zhang2016}.
Optical flow encodes the motion field in a scene and represents the pattern of apparent object motion.
It is a useful motion representation and often acts in a complementary way to spatial features~\cite{Wang2016Temporal, simonyan2014two, yue2015beyond}.

In this paper, we explore the use of optical flow along with spatio-temporal features in a two stream 3-D convolutional architecture to better capture long-range temporal structure in videos.
This framework takes in stacked video frames as well as dense optical flow fields as inputs in two separate streams, and a subsequent convolution operation learns a hierarchical representation of the features.
The two streams are fused at different levels in both proposal generation and activity classification stages with a view to efficiently reuse the computed features for multiple tasks in the detection pipeline.

One of the major problems of `proposal based detection' is that it can suffer from an extreme foreground-background (positive-negative) class imbalance during training. The foreground action proposals (positives) typically account for only a tiny fraction of all possible segments which includes a large number of background segments (negatives).
\textcolor{black}{Though using a separate proposal stage reduces the number of candidate segments by filtering out most background samples, still the classification stage typically has to evaluate hundreds of such segments where only a few are actual foreground activities~\cite{Lin2017}.}
\textcolor{black}{Such an imbalance causes two types of problems for a detection architecture.}
First, training is inefficient as loss is computed and back-propagated for a large number of relatively unimportant candidate proposals.
Second, a large portion of negative examples is easy and can adversely affect the training as these examples contribute no useful learning.
The problem is mainly addressed by maintaining a manageable balance between positive and negative training examples via sampling heuristics (fixed positive-negative ratio of 1:2)~\cite{ren2015faster}, hard negative mining~\cite{Shrivastava2016}, bootstrapping~\cite{Sung_1996} \textit{etc}.
A recent work~\cite{Lin2017} uses dynamically scaled focal loss where the scaling factor down-weighs the contribution of the easy examples during training and rapidly focuses on hard examples.

Inspired by the success of the Online Hard Example Mining (OHEM)~\cite{Shrivastava2016} for fast R-CNN~\cite{girshick2015fast} object detection, we employ a similar strategy where training proposals are subsampled according to their chances of being misclassified.
OHEM not only boosts the detection performance significantly but also is computationally efficient as the hard examples are mined online.
OHEM involves only forward pass operation through the R-C3D network for all the generated candidate proposals.
Then, instead of heuristically sampling the proposals, they are ranked according to the classification and localization loss values and only the top few (\textit{i.e.}, the worst performers) are selected.
Training is performed by back-propagating errors only for these chosen few hard examples.
This improves the performance of the model since good proposals (hard examples)
are selected for updating the model instead of randomly sampled ones.

We perform extensive comparisons of \modelname to state-of-the-art activity detection methods using three publicly available benchmark datasets - THUMOS'14~\cite{THUMOS14}, ActivityNet~\cite{caba2015activitynet} and Charades~\cite{sigurdsson2016hollywood}.
New state-of-the-art results are achieved on THUMOS'14 and Charades, while the detection performance on ActivityNet is competitive.
A preliminary version of this work was published in~\cite{Xu2017}.
This paper additionally explores an optical flow based two-stream architecture, which better captures the motion of objects present in the scene, resulting in better detection of the activities.
We also make training more robust by addressing the class imbalance problem with an online hard mining strategy.
We perform a detailed ablation analysis of the optical flow and OHEM with the basic R-C3D architecture~\cite{Xu2017} and show their effectiveness for efficient temporal activity detection in untrimmed videos.

To summarize, the main contributions of our paper are:
\begin{itemize}[noitemsep,nolistsep]
    \item a jointly optimized activity detection model with combined activity proposal and classification stages that can detect variable length activities;
    \item efficient integration of flow based motion stream and online hard example mining to spatio-temporal features from stacked video frames to boost activity detection performance in untrimmed videos;
    \item fast detection speeds achieved by sharing fully-convolutional C3D features between the proposal generation and classification parts of the network.
\end{itemize}

The rest of the paper is organized as follows.
Section \ref{sec:related_work} gives a description of the state-of-the-art approaches in activity detection, optical flow computation and hard example mining for robust training.
The activity detection approach, including the convolutional feature extraction, proposal generation and classification, optical flow as a second stream and OHEM are described in Section \ref{sec:approach}.
Experimental results and comparisons with state-of-the-art methods are presented in Section \ref{sec:experiments}.
Finally, conclusions are drawn in Section \ref{sec:conclusion}.

\section{Related Work}
\label{sec:related_work}

\subsection{Activity Detection}
There is a long history of activity recognition, or classifying trimmed video clips into fixed set of categories~\cite{ji20133d, laptev2008learning, yue2015beyond, simonyan2014two, wang2013action, Zheng2016}.
Activity \textit{detection} in untrimmed videos, on the other hand, has emerged as a new challenging problem over the past few years and is more practical as most real-life videos are unsegmented and contain multiple activities.

Activity detection can be broadly categorized into two types: spatio-temporal activity detection and temporal activity detection.
Spatio-temporal activity detection~\cite{Weinzaepfel_2015_ICCV, Yu_2015_CVPR} aims to localize spatiotemporal action tubes over consecutive video frames while temporal activity detection~\cite{Gaidon2013} predicts the start and end times of the activities within untrimmed long videos.
Spatio-temporal activity detection tasks tend to be computationally heavy and require more effort in annotating the training data than temporal detection.
In this work, we only focus on supervised temporal activity detection.

\textbf{Sliding-window methods.} Prior to this work, existing temporal activity detection approaches were dominated by models that use sliding temporal windows to generate segments and subsequently classify them with activity classifiers trained on multiple features~\cite{karaman2014fast, oneata2014lear, shou2016temporal, wang2014action}. Sliding windows can be thought of as a primitive method to generate temporal proposals for actions.
Most of these methods have stage-wise pipelines which are not trained jointly, and therefore have limited ability to recover from errors accumulated in each stage.
Moreover, the use of exhaustive sliding windows is computationally inefficient and constrains the boundary of the detected activities by the sliding windows' duration and strides. 

\textbf{Frame/Snippet-level methods.} Recently, several approaches have leveraged recurrent networks to avoid exhaustive sliding window search in detecting activities with variable lengths.
\cite{escorcia2016daps, ma2016learning, montes2016temporal, Singh2016a, yeung2016end} model the temporal evolution of activities using recurrent neural nets (RNNs) or long-short term memory (LSTM) based networks and predict an activity (or background) label for each frame. These frame-level labels are then merged into variable-length segments.
The deep action proposal model~\cite{escorcia2016daps} uses LSTM to encode C3D features of every 16-frame video chunk, and directly regresses and classifies activity segments without an extra proposal generation stage.
\cite{sstad_buch_bmvc17} extends the action proposal model in~\cite{escorcia2016daps} to design a single-pass network for end-to-end temporal action detection that  directly outputs the temporal bounds and corresponding action classes for the detections. 
More recently, CDC~\cite{shou2017cdc} and SSN~\cite{Zhao2017} propose bottom-up activity detection by first predicting at the frame-level/snippet-level and then fusing these predictions.

\textbf{Proposal-classifier methods.}
In this work, we introduce a two stage approach to activity detection: the first, ``proposal'' stage generates temporal proposals containing non-background activity, and the subsequent ``classifier'' stage applies a classifier to each proposal to obtain the detected activity.
We avoid recurrent layers, instead encode a large video buffer with a fully-convolutional 3D ConvNet. We use 3D RoI pooling, inspired by RoI pooling in Faster R-CNN~\cite{ren2015faster} to allow feature extraction from variable length proposals.
3D RoI pooling divides the 3D video encoding into 3D bins and samples values from there.
A contemporaneous method for spatio-temporal activity detection, Tube-CNN~\cite{hou2017tube}, also employs a proposal classification appraoch.
They first divide the video into equal-length clips and and generate a set of tube proposals for each clip. Then, the tube proposals are linked together to perform spatio-temporal action detection.
Tube-CNN~\cite{hou2017tube} proposes tube of interest (ToI) pooling to extract features from tube proposals, which consist of a fixed number of bounding boxes of variable sizes.
The ToI pooling performs RoI pooling in each of the bounding boxes, followed by a temporal pooling over the feature encoding of all the bounding boxes.
\textcolor{black}{Contrary to the 3D RoI pooling proposed in this paper, which extracts features of the temporal proposals from the whole video feature encoding, the tube of interest pooling divides the 3-D pooling into a sequence of 2-D spatial pooling and then 1-D temporal pooling.}

\textbf{Weakly-supervised methods.} Aside from supervised activity detection, a recent work~\cite{Wang2017UntrimmedNets} has addressed weakly supervised activity localization from data labeled only with video level class labels by learning attention weights on shot based or uniformly sampled proposals.
Another type of weakly supervised temporal activity localization is provided with an action sequence and paired video without temporal annotation that requires to localize each action to the input video under the action sequential constraint. The model in~\cite{Richard_2016_CVPR} addresses this problem by a dynamic programming based strategy.
We only focus on supervised temporal activity localization in this paper where the ground truth temporal annotation for each activity is provided.

\subsection{Object Detection}
Activity detection in untrimmed videos is closely related to object detection in images.
The inspiration for our work, Faster R-CNN~\cite{ren2015faster}, extends R-CNN~\cite{girshick2014rich} and Fast R-CNN~\cite{girshick2015fast} object detection approaches, incorporating RoI pooling and a region proposal network.
Compared to recent object detection models \textit{e.g.}, SSD~\cite{liu2016ssd} and R-FCN~\cite{li2016r}, Faster R-CNN is a general and robust object detection framework that has been deployed on different datasets with little data augmentation effort. 
Like Faster R-CNN, our \modelname model is also designed with the goal of easy deployment on varied activity detection datasets.
It avoids making certain assumptions based on unique characteristics of a dataset, such as the UPC model for ActivityNet~\cite{montes2016temporal} which assumes that each video contains a single activity class.

\subsection{Optical Flow}
Over the last few years, use of optical flow to represent motion has improved many video related tasks such as activity classification~\cite{simonyan2014two, Wang2016Temporal}, video description~\cite{Venugopalan2015}, visual odometry~\cite{Agrawal2015} and pose estimation in videos~\cite{Pfister2015} among others.
Flow has been used as additional input in detecting activities too.
Histogram of Optical Flow was used in~\cite{caba2016fast} to characterize the temporal signature of the actions and to generate activity proposals by learning a sparse dictionary of the features.
We, on the other hand, use optical flow as an additional input in a `classification by proposal' framework for temporal activity detection.
Concurrent to our work, Structured Segment Network~\cite{Zhao2017} models each activity instance as a composition of three major stages namely `starting', `course' and `ending', and uses both RGB values and optical flow field features for each stage to detect activities.
In a similar framework as above, Dai \textit{et. al.}~\cite{Dai2017}, also uses optical flow and RGB features from the candidate proposals as well as the surrounding contexts and has shown good activity detection performance in benchmark datasets.
In the temporal activity detection models~\cite{Zhao2017,Dai2017}, the two-stream action recognition model~\cite{simonyan2014two} with optical flow branch is used as feature extractor where 2D convolutional operations are applied on the optical flow input stream. However, 3D convolutional operations are applied on the optical flow input stream in our case.
We also employ online hard mining of training examples to further boost the activity detection performance.
Next, we will discuss, in brief, about some of the relevant works in hard mining of training examples.

\begin{figure*}[ht]
\begin{center}
\includegraphics[width=\linewidth]{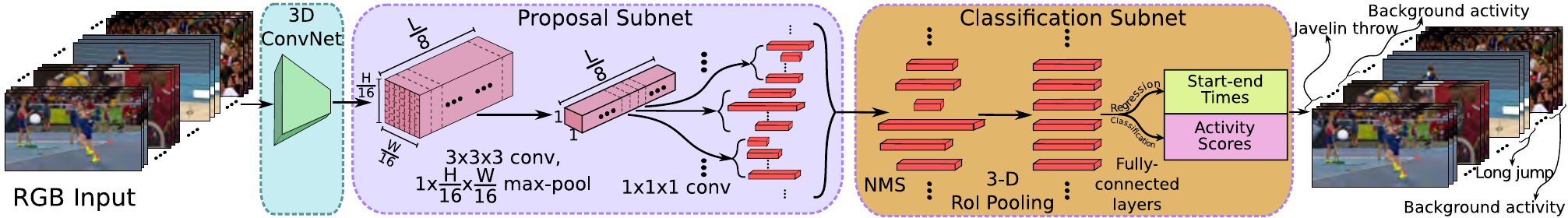}
\end{center}
\vskip -0.15in
\caption{Single stream \modelname model architecture with only RGB frames as input.
The 3D ConvNet takes raw video frames as input and computes convolutional features.
These are input to the Proposal Subnet that proposes candidate activities of variable length along with confidence scores.
The Classification Subnet filters the proposals, pools fixed size features and then predicts activity labels along with refined segment boundaries.}
\label{fig:architecture}
\end{figure*}

\subsection{Hard Mining}
Class imbalance between positive and negative proposals is one of the main obstacles for training a jointly optimized object or activity detection system.
One of the earliest attempts~\cite{Sung_1996} tried to address this problem by freezing the model during training and letting it choose the examples which are hard depending on the misclassification errors.
At the next iteration the model was trained with these hard examples and it became more and more robust until convergence.
The deformable part based object detection model~\cite{Felzenszwalb2010Object} uses a slight variation of the above, where easily classified examples are removed along with the addition of hard examples.
Some of the recent works~\cite{Simo2014, Loshchilov2015} select hard examples while training deep networks based on the current loss for each example.
Lin \textit{et. al.}~\cite{Lin2017} proposes a new loss function for dealing with class imbalance where the loss for hard examples is dynamically set to high values until confidence in the correct class increases.
Region-Based Object Detectors with OHEM~\cite{Shrivastava2016}, on the other hand, proposes an online hard example mining strategy where the detection network is used in the inference mode to get a list of proposals ranked in descending order of their loss values.
The training set is formed of the top few hard examples from this list and these are used for back-propagating the loss through the network.
The boost in performance for object detection and the online nature of the algorithm inspire us to adopt OHEM for the R-C3D network.

\section{Approach}
\label{sec:approach}

In this section, first, the single stream \textit{Region Convolutional 3D Network (R-C3D)}, a convolutional neural network for activity detection in continuous video streams will be described.
Next, the two stream architecture with OHEM will be discussed.
The single stream network, illustrated in Figure~\ref{fig:architecture}, consists of three components: a shared 3D ConvNet feature extractor~\cite{tran2015learning}, a temporal proposal stage, and an activity classification and refinement stage.
To enable efficient computation and joint training, the proposal and classification sub-networks share the same C3D feature maps.
The proposal subnet predicts variable length temporal segments that potentially contain activities, while the classification subnet classifies these proposals into specific activity categories or background, and further refines the proposal segment boundaries.
A key innovation is to extend the 2D RoI pooling in Faster R-CNN to 3D RoI pooling which allows our model to extract features at various resolutions for variable length proposals.
Next, we describe different parts of the architecture starting with the shared video feature hierarchies in Sec.~\ref{sec:feature}.

\subsection{3D Convolutional Feature Hierarchies}
\label{sec:feature}

We use a 3D ConvNet to extract rich spatio-temporal feature hierarchies from a given input video buffer.
It has been shown that both spatial and temporal features are important for representing videos, and a 3D ConvNet encodes rich spatial and temporal features in a hierarchical manner.
The input video frames have dimension $3\times L\times H\times W$, where 3 is the number of color channels, $L$ is the number of frames, $H$ is the height and $W$ is the width.
The architecture of the 3D ConvNet is taken from the C3D architecture proposed in~\cite{tran2015learning}.
However, unlike~\cite{tran2015learning}, the input to our model is of variable length.
We adopt the convolutional layers (\texttt{conv1a} to \texttt{conv5b}) of C3D, so that a feature map $C_{conv5b}\in \mathbb{R}^{512\times \frac{L}{8} \times \frac{H}{16}\times \frac{W}{16}}$ ($512$ is the channel dimension of the layer \texttt{conv5b}) is produced as the output of this subnetwork.
We use $C_{conv5b}$ activations as the shared input to the proposal and classification subnets.
The height ($H$) and width ($W$) of the frames are taken as 112 each following~\cite{tran2015learning}.
The number of frames $L$ can be variable and is only limited by memory.
Note that the same frames are used for computing optical flow fields in the second stream.

\subsection{Temporal Proposal Subnet}
\label{sec:proposal}

To allow the model to predict variable length proposals, we incorporate anchor segments into the temporal proposal sub-network.
The subnet predicts potential proposal segments with respect to anchor segments and a binary label indicating whether the predicted proposal contains an activity or not.
The anchor segments are pre-defined multiscale windows centered at $L/8$ uniformly distributed temporal locations.
\textcolor{black}{Each temporal location specifies $K$ anchor segments, each at different scales.}
Thus, the total number of anchor segments is $(L/8) * K$.
The same set of $K$ anchor segments exists in different temporal locations, which ensures that the proposal prediction is temporally invariant.
The anchors serve as reference activity segments for proposals at each temporal location, where the maximum number of scales $K$ is dataset dependent.

To obtain features at each temporal location for predicting proposals, we first add a 3D convolutional filter with kernel size $3\!\times \!3\!\times \!3$ on top of $C_{conv5b}$ to extend the temporal receptive field for the temporal proposal subnet.
Then, we downsample the spatial dimensions (from $\frac{H}{16}\!\times \!\frac{W}{16}$ to $1\!\times \!1$) to produce a \textit{temporal only} feature map $C_{tpn}\in \mathbb{R}^{512\times \frac{L}{8} \times 1\times 1}$ by applying a 3D max-pooling filter with kernel size $1\!\times \!\frac{H}{16} \!\!\times \!\!\frac{W}{16}$.
The 512-dimensional feature vector at each temporal location in $C_{tpn}$ is used to predict a relative offset $\{\delta c_{i},\delta l_{i}\}$ to the center location and the length of each anchor segment $\{c_{i},l_{i}\},i\in \{1,\cdots,K\}$.
It also predicts the binary scores for each proposal being an activity or background.
The proposal offsets and scores are predicted by adding two $1\!\times \!1\!\times \!\!1$ convolutional layers on top of $C_{tpn}$.

\noindent{\textbf{Training}:} For training, we need to assign positive/negative labels to the anchor segments.
Following the standard practice in object detection~\cite{ren2015faster}, we choose a positive label if the anchor segment 1) overlaps with some ground-truth activity with temporal Intersection-over-Union (tIoU) higher than 0.7, or 2) has the highest tIoU overlap with some ground-truth activity.
If the anchor segment has tIoU overlap lower than 0.3 with all ground-truth activities, then it is given a negative label.
All others are held out from training.
For proposal regression, ground truth activity segments are transformed with respect to nearby positive anchor segments using the coordinate transformations described in Sec.~\ref{sec:optimization}.
Generally, the number of negative proposals is much more than the positive proposals due to smaller number of ground truth activity segments.
To avoid producing a degenerate proposal generation module by the presence of overwhelming number of negative candidates, we fixed the foreground-to-background ratio to $1\!:\!1$ per training batch, and the batch size in the proposal subnet is set to be 64.

\subsection{Activity Classification Subnet}
\label{sec:classification}

The activity classification stage has three main functions: 1) selecting proposal segments from the previous stage, 2) three-dimensional region of interest (3D RoI) pooling to extract fixed-size features for selected proposals, and 3) activity classification and boundary regression for the selected proposals based on the pooled features.

Some activity proposals generated by the proposal subnet highly overlap with each other and some have low proposal scores indicating low confidence.
Following the standard practice in object detection~\cite{Felzenszwalb2010Object, ren2015faster} and activity detection~\cite{shou2016temporal, yeung2016end}, we employ a greedy Non-Maximum Suppression (NMS) strategy to eliminate highly overlapping and low confidence proposals.
The NMS threshold is set to 0.7.

The selected proposals can be of variable length.
However we need to extract fixed-size features for each of them in order to use fully connected layers for further activity classification and regression.
We design a 3D RoI pooling layer to extract the fixed-size volume features for each variable-length proposal from the shared convolutional features $C_{conv5b} \in \mathbb{R}^{512\times (L/8) \times 7\times 7}$ (shared with the temporal proposal subnet).
Specifically, in 3D RoI pooling, an input feature volume of size, say, $ l \!\times\! h \!\times\! w$ is divided into $ l_s \!\times \!h_s\!\times \!w_s$ sub-volumes each with approximate size $ \frac{l}{l_s} \!\times\! \frac{h}{h_s} \!\times\! \frac{w}{w_s}$, and then max pooling is performed inside each sub-volume.
In our case, suppose a proposal has the feature volume of size $ l_p \!\times\! 7 \!\times\! 7$ in $C_{conv5b}$, then this feature volume will be divided into $1 \!\times\! 4 \!\times\! 4$ grids and max pooled inside each grid.
Thus, proposals of variable lengths give rise to  output volume features of the same size $512 \!\times\! 1 \!\times\! 4 \!\times\! 4$.

The output of the 3D RoI pooling is fed to a series of two fully connected layers.
Here, the proposals are classified to activity categories by a classification layer and the refined start-end times for these proposals are given by a regression layer.
The classification and regression layers are also two separate fully connected layers and for both of them the input comes from the aforementioned fully connected layers (after the 3D RoI pooling layer).

\noindent\textbf{Training:} We need to assign activity label to each proposal for training.
An activity label is assigned if the proposal has the highest tIoU overlap with a ground-truth activity, and at the same time, the tIoU overlap is greater than 0.5.
A background label is assigned to proposals with tIoU overlap lower than 0.5 with all ground-truth activities.
The batch size in the classification subnet is set as 128, and training batches are chosen with positive/negative ratio of $1\!\!:\!\!3$.
Later we will show that substituting this heuristic with OHEM increases the efficiency of the detection further in Sec.~\ref{sec:OHEM}.

\subsection{Optimization}
\label{sec:optimization}

We train the network by optimizing both the classification and regression tasks jointly for the two subnets.
The softmax loss function is used for classification, and smooth L1 loss function~\cite{girshick2015fast} is used for regression.
Specifically, the objective function is given by:
\small
\begin{equation}
    \hspace{-2mm} Loss = \frac{1}{N_{cls}} \sum\limits_i L_{cls} (a_i, a_i^*) + \lambda \frac{1}{N_{reg}} \sum\limits_i a_i^* L_{reg} (t_i, t_i^*)
\vspace{-2mm}
\label{eq:loss}
\end{equation}
\normalsize

where $N_{cls}$ and $N_{reg}$ stand for batch size and the number of positive anchor/proposal segments, $\lambda$ is the loss trade-off parameter and is set to a value $1$.
$i$ is the anchor/proposal segments index in a batch, $a_i$ is the predicted probability of the proposal or activities, $a_i^*$ is the ground truth, $t_i = \{\delta \hat{c}_i,\delta \hat{l}_i\}$ represents predicted relative offset to transformed ground truth segments. 
$t_i^* = \{\delta c_i,\delta l_i \}$ represents the coordinate transformation of ground truth segments to anchor segments or proposals. The coordinate transformations are computed as follows:
\begin{equation}
\vspace{-0.1in}
\left\{
\begin{gathered}
  \delta c_i = (c_i^* - c_i) / l_i \\
  \delta l_i = log(l_i^* / l_i)
\end{gathered}
\right.
\label{eq:transformation}
\end{equation}
where $c_i$ and $l_i$ are the center location and the length of anchor segments or proposals while $c_i^*$ and $l_i^*$ denote the same for the ground truth activity segments.

In our \modelname model, the above loss function is applied for both the temporal proposal subnet and the activity classification subnet.
In the proposal subnet, the binary classification loss $L_{cls}$ predicts whether the proposal contains an activity or not, and the regression loss $L_{reg}$ optimizes the relative offset between proposals and ground truths.
Here the losses are activity class agnostic.
For the activity classification subnet, the multiclass classification loss $L_{cls}$ predicts the specific activity class for the proposal, and the number of classes are the number of activities plus one for the background.
The regression loss $L_{reg}$ optimizes the relative displacement between activities and ground truths.
All four losses for the two subnets are optimized jointly.
The network is optimized using SGD solver with momentum 0.9 and weight decay 0.0005.

\subsection{Two Stream Model}
\label{sec:optflow}

\begin{figure*}[ht]
\begin{center}
\includegraphics[width=\linewidth]{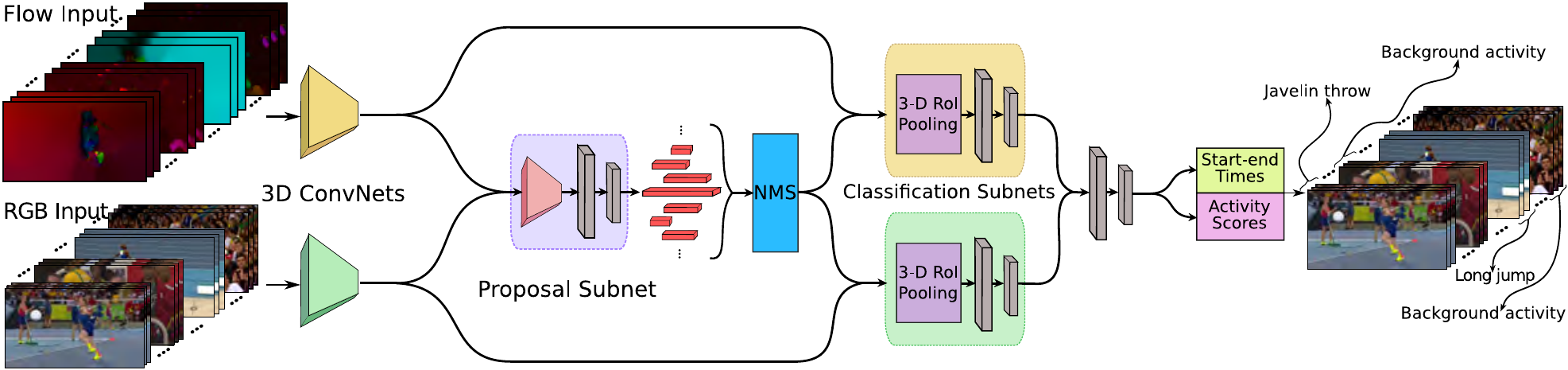}
\end{center}
\vskip -0.15in
\caption{Two stream \modelname architecture.
The top stream receives the stacked optical flow fields as input while the bottom stream receives the RGB frames as input.
Two separate 3D convnets operate on the two inputs to produce the respective feature maps.
These two feature maps are fused and fed to the proposal subnet which proposes candidate activity proposals along with their confidence scores.
The classification subnet works on features from two separate streams but on the same set of proposals.
The outputs of the two classification subnets are fused for the final activity classification and start-end time regression.
}
\label{fig:architecture_two_stream}
\end{figure*}

Action representation is crucial for good performance.
Inspired by the recent success of two-stream CNNs for action classification~\cite{simonyan2014two}, we posit that a natural approach for better action representation is to use optical flow features in a second stream.
The input is a stack of optical flow fields computed for each frame of the first stream (denoted as the RGB stream). 
The two-stream \modelname architecture is shown in Fig.~\ref{fig:architecture_two_stream}.

A dense optical flow field for two consecutive frames of a video gives a two-dimensional vector at each pixel. The horizontal and vertical components of the flow are taken as two separate channels.
TVL1 optical flow algorithm~\cite{zach2007duality} is used to compute the optical flow fields.
The optical flow input has dimension $2\times (L-1) \times H \times W$. 
A 3-D convolutional operation is performed separately on this stream.
The output conv feature maps from both the streams are fused before the proposal generation stage.
The fusion of the two features allows us to use the same proposal subnet (Sec. \ref{sec:proposal}) without having to learn too many extra parameters.
We tried two different fusion strategies.
In the first one, the output of the two streams after the 3D convolution stage is element-wise summed, while in the second, the feature maps from the optical flow stream are added as additional channels to the RGB stream feature maps.
The fused features from the two streams are used to generate the proposals using the Proposal Subnet (Sec. \ref{sec:proposal}).

After the proposals are obtained, NMS is performed to suppress the highly overlapping and low confidence proposals as described in Sec. \ref{sec:classification}.
The proposals that remain after the NMS step are projected separately on two \emph{conv5b} feature maps obtained in two different streams.
3-D RoI pooling is performed on the projected features in both streams separately.
The RoI pooled proposal features are then passed through a series of two fully-connected layers.
Before performing the final classification and boundary regression of the candidate proposals, the outputs from the two separate streams are fused.
Similar to the Proposal Subnet, here also we explore two different fusion strategies.
Firstly, we sum the response of the two streams element-wise and secondly the two responses are concatenated to be fed to the final activity classification and boundary regression layers.
We show results for both fusion strategies unless otherwise mentioned.

Since the final classification or regression are performed on the fused feature vector, the loss formulation for the joint training remains the same as the single stream architecture (Sec. \ref{sec:optimization}).
This is also true for the proposal classification and time regression losses in the Proposal Subnet.

\subsection{Online Hard Example Mining}
\label{sec:OHEM}
Mining hard examples is aimed at choosing better, more informative examples for training the model.
Inspired by the effectiveness as well as easy integration of the online hard example mining strategy in Fast R-CNN image detection pipeline~\cite{Shrivastava2016}, we experiment with a similar strategy in our R-C3D network.
While \modelname with OHEM has a different strategy for choosing training examples than the original R-C3D, they follow the same prediction procedure.
The original \modelname chooses training examples using a fixed positive-to-negative example ratio in fixed sized batches.
\modelname with OHEM precomputes the loss for all of the candidate proposals and then chooses only the hard examples, i.e. the ones with high loss.
Hard training examples could have been mined for both the proposal and the classification subnets.
However, the proposal subnet involves binary classification and thus the training examples are more evenly balanced between positive and negative classes.
In this paper, we apply OHEM only in the classification subnet as it involves multi-class classification and thus suffers more from training data imbalance when the number of categories is large.

In the classification subnet, we add an extra read-only classification branch which shares the weights with the original classification subnet.
The ``read-only'' classification branch is only used to compute loss for all the proposals generated by the proposal subnet and does not update its weights during the backpropagation stage.
Specifically, the sum of the classification loss and regression loss for each proposal is computed in this read-only clone.
The loss values represent how well the current network performs on each proposal.
Proposals are sorted according to the sum of the losses in descending order and only top 128 are taken to form the training mini-batch.
\textcolor{black}{After the hard examples are chosen, the backward pass is performed only with these hard examples in the original classification subnet.}
Since the read-only classification subnet shares weights with the original classification subnet, it also gets updated with the same weights in the next iteration and the losses are computed using the new weights.

\subsection{Prediction}
\label{sec:prediction}

Activity prediction in R-C3D consists of two steps.
First, the proposal subnet generates candidate proposals and predicts the start-end time offsets as well as proposal score for each.
Then the proposals are refined via NMS with threshold value 0.7.
After NMS, the selected proposals are fed to the classification network to be classified into specific activity classes, and the activity boundaries of the predicted proposals are further refined by the regression layer.
The boundary prediction in both proposal subnet and classification subnet is in the form of relative displacement of center point and length of segments.
In order to get the start time and end time of the predicted proposals or activities, inverse coordinate transformation to Equation \ref{eq:transformation} is performed.

\modelname accepts variable length input videos.
However, to take advantage of the vectorized implementation in fast deep learning libraries, we pad the last few frames of short videos with last frame, and break long videos into buffers (limited by memory only).
NMS at a lower threshold (0.1 less than the mAP evaluation threshold) is applied to the predicted activities to get the final activity predictions.
\section{Experiments}
\label{sec:experiments}

We evaluate \modelname on three large-scale activity detection datasets - THUMOS'14~\cite{THUMOS14}, Charades~\cite{sigurdsson2016hollywood} and ActivityNet~\cite{caba2015activitynet}.
In evaluating the proposals, we follow the evaluation paradigm for the temporal localization task in ActivityNet dataset (ref. ~\ref{exp:activitynet}) and use the Area Under the AR vs AN curve (AUC) at 100 proposals per video which is averaged across ten different tIoU thresholds uniformly distributed between 0.5 and 0.95.
Activity detection results are shown in terms of mean Average Precision - mAP@$\alpha$ where $\alpha$ denotes different tIoU thresholds, as is the common practice in the literature.
Section~\ref{exp:speed} provides the detection speed in comparison to state-of-the-art activity detection approaches.

\subsection{Experiments on THUMOS'14}
\label{exp:thumos14}

THUMOS'14 activity detection dataset contains over 24 hours of video from 20 different sport activities.
The training set contains 2765 trimmed videos while the validation and the test sets contain 200 and 213 untrimmed videos respectively.
This dataset is particularly challenging as it consists of very long videos (up to a few hundreds of seconds) with multiple activity instances of very small duration (up to few tens of seconds).
Most videos contain multiple activity instances of the same activity class.
In addition, some videos contain activity segments from different classes.

\noindent{\textbf{Experimental Setup}:}
We divide 200 untrimmed videos from the validation set into 180 training and 20 held out videos to get the best hyperparameter setting.
All 200 videos are used as the training set and the final results are reported on 213 test videos.
Since the GPU memory is limited, we first create a buffer of 768 frames at 25 frames per second (fps) which means approximately 30 seconds of video.
Our choice is motivated by the fact that 99.5\% of all activity segments in the validation set (used here as the training set) are less than 30 seconds long.
These buffers act as inputs to both streams of R-C3D.
We can create the buffer by sliding from the beginning of the video to the end, denoted as the `one-way buffer'.
An additional pass from the end of the video to the beginning is used to increase the amount of training data, denoted as the `two-way buffer'.

We initialize the 3D ConvNet part of our model with C3D weights trained on Sports-1M and finetuned on UCF101 released by the authors in~\cite{tran2015learning}.
We allow all the layers of \modelname to be trained on THUMOS'14 with a fixed learning rate of 0.0001.
The number of anchor segments $K$ chosen for this dataset is 10 with specific scale values [2, 4, 5, 6, 8, 9, 10, 12, 14, 16].
The values are chosen according to the distribution of the activity durations in the training set.
At 25 fps and temporal pooling factor of 8 ($C_{tpn}$ downsamples the input by 8 temporally), the anchor segments correspond to segments of duration  between 0.64 and 5.12 seconds\footnote{$2*8/25=0.64$ and $16*8/25=5.12$}.
Note that, the predicted proposals or activities are relative to the anchor segments but not limited to the anchor boundaries, enabling our model to detect flexible-length activities.

For the two-stream R-C3D, we explored two ways of fusing the RGB and flow streams.
The first strategy is denoted as `concat' and it concatenates both streams before the proposal subnet and before the final classification stage.
The second strategy sums the two streams element-wise and is denoted as `sum'.
We still use the RGB C3D weights released in~\cite{tran2015learning} to initialize the RGB stream.
In order to initialize the flow stream, we follow the pipeline proposed in~\cite{Wang2016Temporal} to get a set of pretrained flow C3D weights on UCF101 flow images.
Before finetuning flow C3D weights on UCF101 flow images, the flow C3D model is initialized with RGB C3D weights.
The flow C3D have the same architecture as that of the RGB C3D network in all the layers except the first layer due to different channel dimensions for the flow and RGB inputs (2 \textit{vs} 3).
To make use of the weights in the first layer of RGB C3D model, we average the weights across channels in first convolution layer of RGB C3D model, and replicate this average weights in the two channels of the optical flow input.
We test the efficiency of the C3D architecture for both the RGB as well as the flow streams.
The RGB C3D shows around 80\% activity recognition accuracy on UCF101 split-1 while the same for the flow C3D is around 70\%.
The learning strategy in flow stream is kept same as the RGB stream.

We apply OHEM on both the RGB single-stream \modelname and two-stream R-C3D.
These two experimental settings are denoted as `Single-stream R-C3D + OHEM' and `Two-stream R-C3D (Sum) + OHEM' respectively.
Note that in the later setting we opt for the element-wise sum strategy of fusion as the performance corresponding to this strategy is better than the concatenation strategy for the Two-stream \modelname without hard mining.
The NMS threshold used for the OHEM setting is $0.7$, and the batch size in the activity classification stage is set as 128 which is the same for the other datasets too.

\begin{table}[!t]
\centering
\caption{Proposal evaluation on THUMOS'14 dataset (in percentage). Average AUC of 100 proposals per video at tIoU thresholds $\alpha \in (0.5,0.95)$ with step 0.05 are reported.} 
\normalsize
\begin{tabular}{l || c} 
\hline
~ & $\alpha \in (0.5,0.95)$  \\ \hline
\specialcellL{Single-stream R-C3D} & 26.36 \\  
\specialcellL{Two-stream R-C3D (Concat)} & 28.75 \\
\specialcellL{Two-stream R-C3D (Sum)} & 29.71 \\ \hline 
\end{tabular}
\label{res:proposal_thumos14}
\end{table}

\begin{table}[!t]
\centering
\caption{Activity detection results on THUMOS'14 (in percentage). mAP at different tIoU thresholds $\alpha$ are reported. Top three performers on THUMOS'14 challenge leaderboard and other results reported in existing papers are shown.
}
\normalsize
\begin{tabulary}{1\linewidth}{l || c c c c c} 
\hline
 ~ & \multicolumn{5}{c}{$\alpha$} \\
 ~ & \!\!0.1  & \!\!0.2  & \!\!0.3  & \!\!0.4 & \!\!0.5 \\ \hline
 \!\!Karaman \textit{et. al}.~\cite{karaman2014fast} & \!\!4.6  & \!\!3.4  &  \!\!2.1 & \!\!1.4 & \!\!0.9 \\ 
 \!\!Wang \textit{et. al}.~\cite{wang2014action} & \!\!18.2  & \!\!17.0  & \!\!14.0 & \!\!11.7 &  \!\!8.3 \\ 
 \!\!Oneata \textit{et. al}.~\cite{oneata2014lear} & \!\!36.6  &  \!\!33.6 & \!\!27.0  & \!\!20.8 & \!\!14.4 \\ 
 \!\!Heilbron \textit{et. al}.~\cite{caba2016fast} & \!\!- & \!\!-  & \!\!-  & \!\!- & \!\!13.5 \\ 
 \!\!Escorcia \textit{et. al}.~\cite{escorcia2016daps} & \!\!- & \!\!-  & \!\!-  & \!\!- & \!\!13.9 \\ 
 \!\!Richard \textit{et. al}.~\cite{Richard_2016_CVPR} & \!\!39.7 & \!\!35.7  & \!\!30.0  & \!\!23.2 & \!\!15.2 \\ 
 \!\!Yeung \textit{et. al}.~\cite{yeung2016end} & \!\!48.9 &  \!\!44.0 &  \!\!36.0 & \!\!26.4 & \!\!17.1 \\ 
 \!\!Yuan \textit{et. al}.~\cite{yuan2016temporal} & \!\!51.4 & \!\!42.6  &  \!\!33.6 & \!\!26.1 & \!\!18.8 \\ 
 \!\!Shou \textit{et. al}.~\cite{shou2016temporal} & \!\!47.7 & \!\!43.5  & \!\!36.3  & \!\!28.7 & \!\!19.0 \\
 \!\!Shou \textit{et. al}.~\cite{shou2017cdc} & \!\!- & \!\!-  & \!\!40.1  & \!\!29.4 & \!\!23.3 \\
 \!\!Dai \textit{et. al}.~\cite{Dai2017} & \!\!- & \!\!-  & \!\!-  & \!\!33.3 & \!\!25.6 \\
 \!\!Zhao \textit{et. al}.~\cite{Zhao2017} & \!\!\textbf{66.0} & \!\!\textbf{59.4}  & \!\!\textbf{51.9}  & \!\!41.0 & \!\!29.8 \\ \hline
 \!\!\specialcellL{Single-stream R-C3D~\cite{Xu2017} \\ (one-way buffer)} \!\!\!& \!\!51.6 & \!\!49.2 & \!\!42.8  & \!\!33.4 & \!\!27.0\\
 \!\!\specialcellL{Single-stream R-C3D~\cite{Xu2017} \\ (two-way buffer)} \!\!\!& \!\!54.5 & \!\!51.5  & \!\!44.8  & \!\!35.6 & \!\!28.9 \\
 \!\!\specialcellL{Two-stream R-C3D \\ (Concat)} \!\!\!& \!\!54.5 & \!\!52.2  & \!\!46.9  & \!\!40.0 & \!\!33.1 \\
 \!\!\specialcellL{Two-stream R-C3D \\ (Sum)} \!\!\!& \!\!56.6 & \!\!54.2  & \!\!48.9  & \!\!40.6 & \!\!33.4 \\
 \!\!\specialcellL{Single-stream R-C3D \\ + OHEM} \!\!\!& \!\!57.4 & \!\!54.9  & \!\!51.1  & \!\!\textbf{43.1} & \!\!35.8 \\ 
 \!\!\specialcellL{Two-stream R-C3D (Sum) \\ + OHEM} \!\!\!& \!\!56.9 & \!\!54.7  & \!\!51.2  & \!\!43.0 & \!\!\textbf{36.1} \\ \hline
\end{tabulary}
\label{tab:res_thumos14}
\end{table}

\begin{table*}[!t]
\centering
\caption{Per-class AP at tIoU threshold $\alpha=0.5$ on THUMOS'14 (in percentage).}
\normalsize
 \begin{tabular}{l || c c c || c c c} 
 \hline
 ~ & \cite{oneata2014lear} & \cite{yeung2016end} &  \cite{shou2016temporal} & \specialcellL{Single-stream R-C3D \\ (two-way buffer)~\cite{Xu2017}} & \specialcellL{Two-stream R-C3D \\ (Sum)} & \specialcellL{Single-stream R-C3D \\ + OHEM} \\ \hline
 Baseball Pitch  & 8.6 & 14.6 & 14.9 &  26.1  & 19.9 & \bf{29.9} \\ 
 Basketball Dunk & 1.0 & 6.3 &20.1 & 54.0 &\bf{55.3} &48.6 \\ 
 Billiards & 2.6 & 9.4 & 7.6 & 8.3 &11.2 &\bf{19.8} \\ 
 Clean and Jerk  & 13.3 & \bf{42.8} &24.8 & 27.9 &33.2 &37.7 \\ 
 Cliff Diving    & 17.7 & 15.6 &27.5 & 49.2 &54.0 &\bf{59.4}  \\ 
 Cricket Bowling & 9.5 & 10.8 & 15.7& 30.6 &31.1 &\bf{32.4} \\ 
 Cricket Shot    & 2.6 & 3.5 &13.8 & 10.9 &11.6 &\bf{18.4} \\ 
 Diving & 4.6    & 10.8 & 17.6 & 26.2 &31.1 &\bf{36.4}  \\ 
 Frisbee Catch   & 1.2 & 10.4 &15.3 & 20.1 &\bf{21.5} &16.9  \\ 
 Golf Swing      & 22.6  & 13.8 &18.2 & 16.1 &32.8 &\bf{42.3} \\ 
 Hammer Throw    & 34.7 & 28.9 &19.1 & 43.2 &\bf{58.3} &57.3 \\ 
 High Jump       & 17.6 & 33.3 &20.0 & 30.9 &\bf{37.9} &37.8 \\ 
 Javelin Throw   & 22.0 & 20.4 &18.2 & 47.0 &47.2 &\bf{59.2} \\ 
 Long Jump       & 47.6 & 39.0 &34.8 & 57.4&62.1 &\bf{63.9}  \\ 
 Pole Vault      & 19.6 & 16.3 &32.1 & 42.7 &\bf{57.7} &57.0 \\ 
 Shotput & 11.9  & 16.6 &12.1 & 19.4 &20.0 &\bf{31.0} \\ 
 Soccer Penalty  & 8.7 & 8.3 &19.2 & 15.8 &19.2 &\bf{22.9} \\ 
 Tennis Swing    & 3.0 & 5.6 &\bf{19.3} & 16.6 &11.6 &12.5 \\ 
 Throw Discus    & 36.2 & 29.5 &24.4 & 29.2 &\bf{41.0} &22.1\\ 
 Volleyball Spiking & 1.4 & 5.2 &4.6 & 5.6 &\bf{11.3} &11.2\\ \hline

 mAP@0.5 &14.4 & 17.1 & 19.0 & 28.9 &33.4 &\bf{35.8} \\ \hline 
 \end{tabular}
\label{tab:per_class_ap}
\end{table*}

\noindent{\textbf{Results}:}
We first evaluate the performance of the temporal proposal subnet.
The proposals are first processed with NMS threshold $0.7$.
The proposal evaluation results in terms of average AUC with 100 proposals per video for single-stream \modelname and two-stream \modelname models are shown in Table~\ref{res:proposal_thumos14}.
The average AUC for single-stream \modelname is 26.36\%.
Both the two-stream extensions improve the proposal evaluation results, with two-stream R-C3D (Concat) having the average AUC 28.75\% and two-stream R-C3D (Sum) 29.71\%.

In Table~\ref{tab:res_thumos14}, we present a comparative evaluation of the activity detection performance of our model with existing state-of-the-art approaches in terms of mAP at tIoU thresholds 0.1-0.5 (denoted as $\alpha$).
From the results corresponding to the single-stream R-C3D, we can see that the mAP@0.5 with the two-way buffer setting is better than the mAP@0.5 with the one-way buffer by 1.9\%.
So we take the two-way buffer setting as the data augmentation strategy for all the two-stream experiments unless otherwise mentioned.

Both the two-stream \modelname achieve better activity detection performance than the single-stream R-C3D.
However the differences in performances between two-stream (sum) and two-stream (concat) are very minor in terms of mAP@$\alpha$ metric, with two-stream (sum) achieving slightly better results.
When OHEM is applied to single-stream \modelname and two-stream \modelname (sum), the performance improves significantly for both the scenarios.
However, the relative improvement of OHEM from two-stream \modelname (sum) is less than the single-stream R-C3D.
With OHEM, single-stream \modelname and two-stream \modelname (sum) have almost the same results.
The two-stream \modelname (sum) architecture with OHEM achieves a new state-of-the-art in the mAP@0.5 metric, which requires a stringent overlap with the ground truth segment.
The absolute improvement is $6.3\%$ over the current state-of-the-art ~\cite{Zhao2017}.

\begin{figure*}[!t]
\centering
\subfigure[THUMOS'14]{
\includegraphics[width=0.9\linewidth]{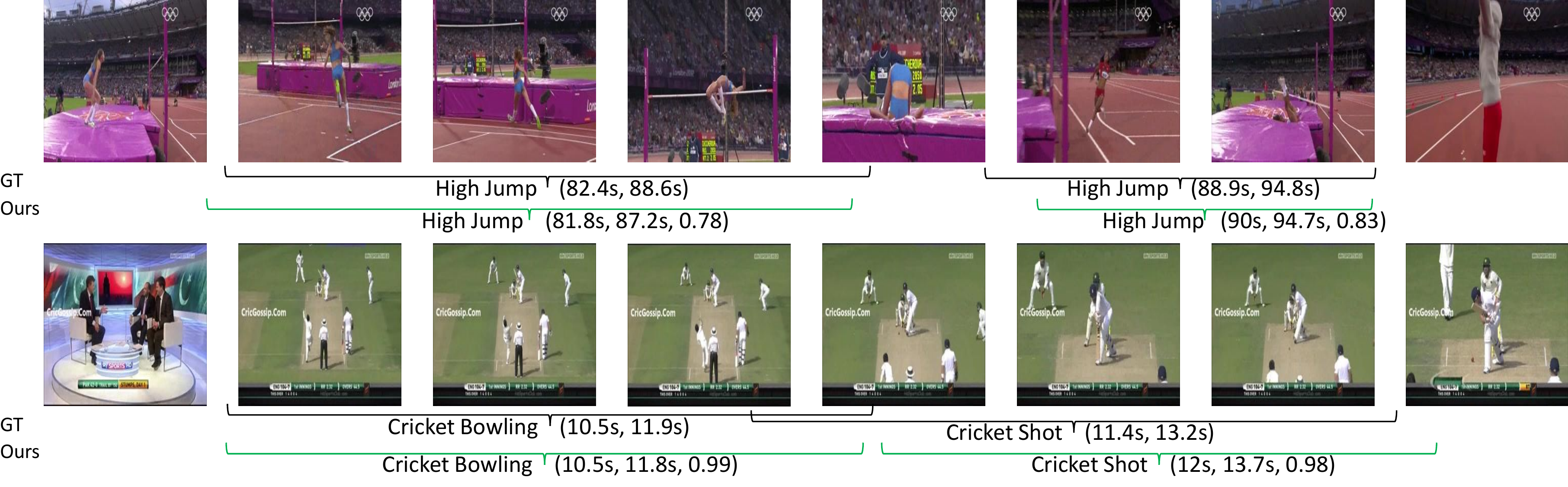}
\label{fig:vis_thumos14}
}
\subfigure[ActivityNet]{
\label{fig:vis_activitynet}
\includegraphics[width=0.9\linewidth]{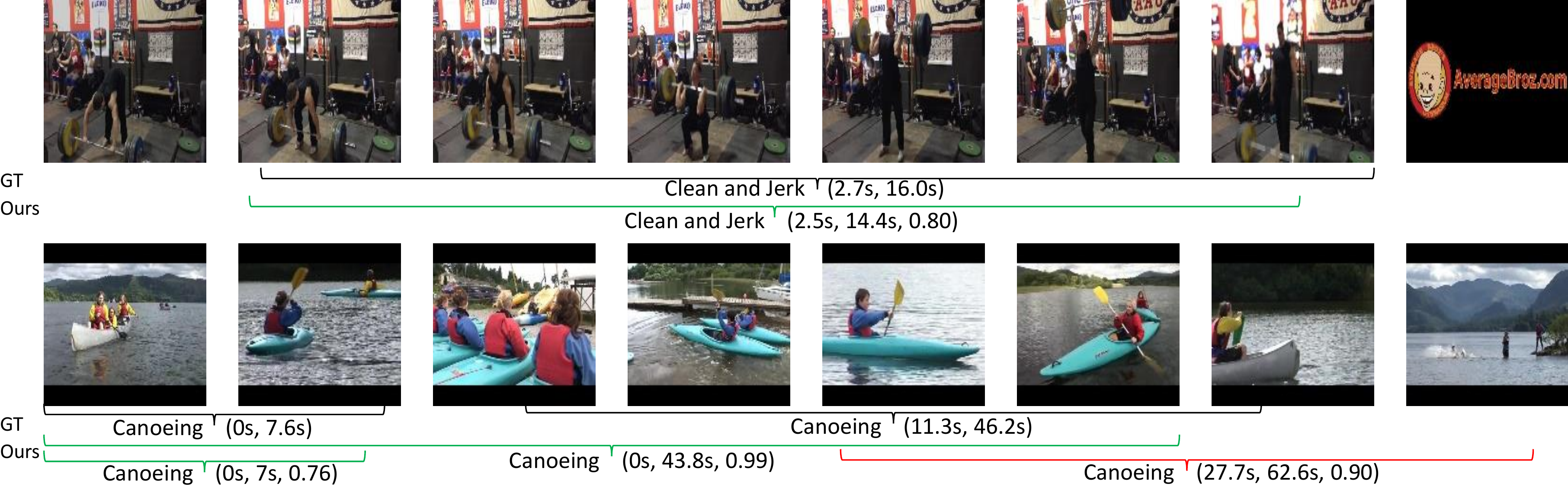}}
\subfigure[Charades]{
\label{fig:vis_charades}
\includegraphics[width=0.9\linewidth]{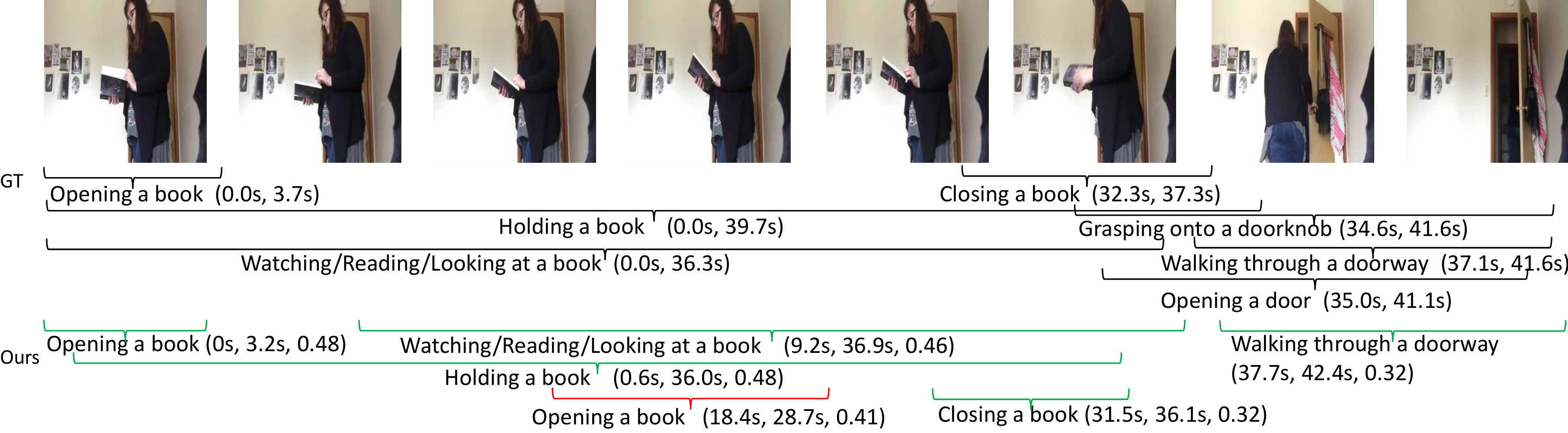}}
\caption
{Qualitative visualization of the predicted activities by single-stream \modelname (best viewed in color). Figure \subref{fig:vis_thumos14} and \subref{fig:vis_activitynet} show results for two videos each on THUMOS'14 and ActivityNet.~\subref{fig:vis_charades} shows the result for one video from Charades. Ground truth activity segments are marked in black. Predicted activity segments are marked in green for correct predictions and in red for wrong ones. Predicted activities with tIoU more than 0.5 are considered as correct. Corresponding start-end times and confidence score are shown inside brackets.}
\label{fig:qualitative}
\end{figure*}

The Average Precision (AP) for each class on THUMOS'14 at tIoU threshold 0.5 is shown in Table~\ref{tab:per_class_ap}.
Compared to other published models, single-stream \modelname outperforms previous methods in most classes and shows significant improvement (by more than 20\% absolute AP over the next best) for activities \textit{e.g.}, Basketball Dunk, Cliff Diving, and Javelin Throw.
With the addition of the optical flow stream, we can see two-stream \modelname (sum) improves the detection performance significantly for activities with obvious motion patterns \textit{e.g.}, Hammer Throw, Golf Swing, Pole Vault and Throw Discus \textit{etc}.
In single-stream \modelname with OHEM, some hard activities with low class precision get further improvement \textit{e.g.}, Billiards, Cricket Shot and Shotput.
Figure~\ref{fig:vis_thumos14} shows some representative qualitative results from two videos on this dataset.

\subsection{Experiments on ActivityNet}
\label{exp:activitynet}

The ActivityNet~\cite{caba2015activitynet} dataset consists of untrimmed videos and is released in three versions.
We use the latest release (1.3) which has 10024, 4926 and 5044 videos containing 200 different types of activities in the train, validation and test sets respectively.
Most videos contain activity instances of a single class covering a great deal of the video.
Compared to THUMOS'14, this is a large-scale dataset both in terms of the number of activities involved and the amount of videos.
Researchers have taken part in the ActivityNet challenge~\cite{activitynetChallenge} held on this dataset.
The performances of the participating teams are evaluated on test videos for which the ground truth annotations are not public.
In addition to evaluating on the validation set, we show our performance on the test set after evaluating it on the challenge server.

\noindent{\textbf{Experimental Setup}:}
Similar to THUMOS'14, the length of the input buffer is set to 768 but, as the videos are long, we sample frames at 3 fps to fit it into the GPU.
This makes the duration of the buffer approximately 256 seconds covering over 99.99\% training activities.
The considerably long activity durations prompt us to set the number of anchor segments $K$ to be as high as 20.
Specifically, we chose the following scales - [1, 2, 3, 4, 5, 6, 7, 8, 10, 12, 14, 16, 20, 24, 28, 32, 40, 48, 56, 64].
Thus the shortest and the longest anchor segments are of durations 2.7 and 170 seconds respectively covering 95.6\% of the training activities.

Considering the vast domain difference of the activities between Sports-1M and ActivityNet, we finetune the Sports-1M pretrained RGB 3D ConvNet model~\cite{tran2015learning} with the training videos of ActivityNet at 3 fps on the activity classification task.
We initialize the RGB 3D ConvNet part of our model with these finetuned weights.
We also use the Sports-1M pretrained RGB 3D ConvNet model to initialize the flow 3D ConvNet model and finetune it on the flow images of the ActivityNet training videos as is done for the THUMOS'14 dataset (ref. Sec.~\ref{exp:thumos14}).
We test the efficiency of the C3D architecture for both the RGB and flow streams on ActivityNet.
The RGB C3D shows around 60\% activity recognition accuracy while the same for the flow C3D is around 32\%.
AcitivityNet being a large scale dataset, the training takes more epochs.
As a speed-efficiency trade-off, we freeze the first two convolutional layers in our model during training.
The learning rate is kept fixed at $10^{-4}$ for first 10 epochs and is decreased to $10^{-5}$ for the last 5 epochs.
Based on the improved results on the THUMOS'14, we choose the two-way buffer setting with horizontal flipping of frames for data augmentation.

\begin{table}[!t]
\centering
\caption{Proposal evaluation on ActivityNet validation set (in percentage). Average AUC of 100 proposals per video at tIoU thresholds $\alpha \in (0.5,0.95)$ with step 0.05 are reported.}
\normalsize
\begin{tabular}{l || c} 
\hline
~ & $\alpha \in (0.5,0.95)$  \\ \hline
\specialcellL{Single-stream R-C3D} & 56.7 \\  
\specialcellL{Two-stream R-C3D (Concat)} & 56.2 \\
\specialcellL{Two-stream R-C3D (Sum)} & 55.6 \\ \hline 
\end{tabular}
\label{res:proposal_act}
\end{table}

\begin{table}[!t]
\centering
\caption{Detection results on ActivityNet in terms of mAP@0.5 (in \%). The top half of the table shows performance from methods using additional handcrafted features while the bottom half shows approaches using deep features only (including ours). Results for~\cite{Singh2016a} are taken from~\cite{activitynetChallenge}.} 
\small
\begin{tabular}{l || c |c c c c} 
\hline
 ~ & train data & validation  & test   \\ \hline
B. Singh  \textit{et. al.}~\cite{Singh2016a} & train+val & -  & 28.8   \\  
G. Singh \textit{et. al}.~\cite{Singh2016b} & train & 34.5  & 36.4   \\
Dai \textit{et. al}.~\cite{Dai2017} & train & 36.2  & 37.5   \\  
\hline
UPC ~\cite{montes2016temporal} & train & 22.5  & 22.3   \\
Zhao \textit{et. al}.~\cite{Zhao2017} & train & -  & 43.3 \\
Single-stream \modelname~\cite{Xu2017} & train & 26.8 & 26.8\\ 
Single-stream \modelname~\cite{Xu2017} & train+val & - & 28.4 \\
Two-stream \modelname (Concat) & train & 25.8 & - \\
Two-stream \modelname (Sum) & train & 26.5 & - \\
Single-stream \modelname + OHEM & train & 27.7 & - \\ \hline
\end{tabular}
\label{res:activitynet}
\end{table}

\begin{table}[!t]
\centering
\caption{Results on ActivityNet in terms of average mAP at tIoU thresholds $\alpha \in (0.5,0.95)$ with step 0.05 (in \%).}
\small
\begin{tabular}{l || c |c c c c} 
\hline
 ~ & train data & validation  & test   \\ \hline
Single-stream \modelname~\cite{Xu2017} & train & 12.7 & 13.1\\ 
Single-stream \modelname~\cite{Xu2017} & train+val & - &  16.7 \\
Single-stream \modelname + OHEM & train & 15.4 & 15.6 \\ \hline
\end{tabular}
\label{res:activitynet_mAP}
\end{table}

\noindent{\textbf{Results}:}
We first evaluate the proposal performance from the proposal subnet after the NMS step with threshold $0.7$.
The proposal evaluation results in terms of average AUC with 100 proposals per video for single-stream \modelname and two-stream \modelname models are shown in Table~\ref{res:proposal_act}.
Both the two-stream extensions do not improve the proposal results and maintain almost the same performance as the single-stream R-C3D.

Table~\ref{res:activitynet} shows mAP@0.5 performance of our models and compares them with the published results from existing activity detection approaches.
In most experiments, the training set is used for training and the performance is shown for either the validation or test data or both.
Some models in Table~\ref{res:activitynet} make use of sophisticated handcrafted features.
The approach in~\cite{Singh2016b} also uses handcrafted motion features like MBH on top of inception and C3D features in addition to dynamic programming based post processing.
Our method is also capable of using other hand engineered features with a possible boost to performance and we keep this as a future task.
UPC is a fair comparison to single stream \modelname as it also uses only C3D features.
However, it relies on a strong assumption that each video on ActivityNet just contains one activity class.
Our approach obtains 4.3\% improvement on the validation set and 4.5\% improvement on the test set over UPC~\cite{montes2016temporal} in terms of mAP@0.5 without such strong assumptions.
When both training and validation sets are used for training, the performance improves further by 1.6\%.

As is the scenario in proposal evaluation, both the two-stream extensions of \modelname provide roughly the same detection performance as the single-stream R-C3D.
One reason can be that the initial flow C3D branch of two-stream \modelname has only around 32\% classification accuracy on the ActivityNet validation set which is significantly lower than the classification performance (60\%) by the initial RGB branch on the same dataset.
As a result of the poor initialization of the flow C3D model, the two-stream extensions do not improve the results much.
Since our proposed \modelname activity detection model is based on low level 3D conv features learned by the C3D models, a better classification model with better initialization can give rise to better activity detection performance.
We expect better performance by initializing with stronger classification architecture like I3D~\cite{Carreira2017} and pre-training with larger and more diverse video dataset like kinetics~\cite{Carreira2017} or Youtube-8M~\cite{Abu2016Youtube}.

We only experiment with OHEM on single-stream R-C3D, since without OHEM, single-stream \modelname performs almost the same as the two-stream \modelname and a single-stream \modelname with OHEM is relatively less expensive in terms of computation.
Single-stream \modelname with OHEM provides better performance with mAP@0.5 reaching $27.7\%$ compared to single-stream \modelname without OHEM.
As further analysis, we divide the videos in the validation set of ActivityNet into three sets (Short/Medium/Long) in equal numbers according to the ground truth activity durations.
Single-stream \modelname with OHEM gets 39.2\% mAP@0.5 on Long set (greater than 53 seconds), 23.2\% mAP@0.5 on Medium set (13 seconds - 53 seconds), and 8.3\% mAP@0.5 on Short set (less than 13 seconds).
This shows that our model is poor on the relatively short activities on this dataset.
The reason might be that for ActivityNet, we sample frames at 3 fps to fit the long videos into the GPU memory.
\textcolor{black}{For short videos, this implies that only a few frames are sampled for proposal feature encoding.}
The poor feature encoding from limited frames might be the reason for the poor performance of these short activities on this dataset.

Our highest mAP@0.5 result is still lower than Zhao \textit{et. al}.~\cite{Zhao2017} which uses an additional set of 200 separately trained binary classifiers to classify whether the predicted activity segments are complete or not.
These completeness classifiers pay special attention to improve the tIoU of the predicted activity segments which is one of the main reasons to boost their detection results. 
Our model can also benefit from using such completeness classifiers at the cost of more computation.

The ActivityNet Challenge in 2017 introduced a new evaluation metric where mAP at 10 evenly distributed thresholds between [0.5, 0.95] are averaged to get the \emph{average mAP}.
We show the performance of single-stream \modelname with and without OHEM in Table~\ref{res:activitynet_mAP}.
Training with videos from the training partition only, the average mAP for the validation and test set come to be 12.7\% and 13.1\% respectively.
If both training and validation data are used during training, the average mAP for the test set increases to 16.7\% showing the benefit of our jointly optimized single-stream model when more data is available for training.
The OHEM extension also improves the single-stream results in terms of average mAP.
Figure~\ref{fig:vis_activitynet} shows some representative qualitative results of \modelname from this dataset.

\subsection{Experiments on Charades}
\label{exp:charades}

Charades~\cite{sigurdsson2016hollywood} is a recently introduced dataset for activity classification and detection.
The dataset consists of 7985 train and 1863 test videos from 157 classes.
The videos are recorded by Amazon Mechanical Turk users based on provided scripts.
Apart from low illumination, diversity and casual nature of the videos containing day-to-day activities, an additional challenge of this dataset is the abundance of overlapping activities, sometimes multiple activities having exactly the same start and end times (typical examples include pairs of activities like `holding a phone' and `playing with a phone' or `holding a towel' and `tidying up a towel').

\noindent{\textbf{Experimental Setup}:}
For this dataset we sample frames at 5 fps, and the input buffer is set to contain 768 frames.
This makes the duration of the buffer approximately 154 seconds covering all the ground truth activity segments in the train set.
As the activity segments are longer, we choose the number of anchor segments $K$ to be 18 with specific scale values [1, 2, 3, 4, 5, 6, 7, 8, 10, 12, 14, 16, 20, 24, 28, 32, 40, 48].
So the shortest anchor segment has a duration of 1.6 seconds and the longest anchor segment has a duration of 76.8 seconds. 
Over 99.96\% of the activities in the training set is under 76.8 seconds.
For this dataset we, additionally, explore slightly different settings of the anchor segment scales, but find that our model is not very sensitive to this hyper-parameter.

We first finetune the Sports-1M pretrained RGB C3D model~\cite{tran2015learning} on the Charades training set at 5 fps and initialize the 3D ConvNet for the RGB stream with these finetuned weights.
The flow stream initialization follows the same pipeline mentioned in Sec.~\ref{exp:thumos14} and Sec.~\ref{exp:activitynet}.
Both the RGB C3D model and flow C3D model have very low activity classification accuracy (9.6\% and 8.8\% respectively), due to the multi-label nature of the activity segments on this dataset.
While training the full model, we freeze the first two convolutional layers in order to accelerate training.
The learning rate is kept fixed at 0.0001 for the first 10 epochs and then decreased to 0.00001 for 5 further epochs.
We augment the data by following the two-way buffer setting and horizontal flipping of frames.

\begin{table}[!t]
\centering
\caption{Proposal evaluation results on Charades (in percentage). Average AUC of 100 proposals per video at tIoU thresholds $\alpha \in (0.5,0.95)$ with step 0.05 are reported.}
\normalsize
\begin{tabular}{l || c} 
\hline
~ & $\alpha \in (0.5,0.95)$  \\ \hline
\specialcellL{Single-stream R-C3D} & 70.0 \\  
\specialcellL{Two-stream R-C3D (Concat)} & 69.2 \\
\specialcellL{Two-stream R-C3D (Sum)} & 69.6 \\ \hline 
\end{tabular}
\label{res:proposal_charades}
\end{table}

\begin{table}[!t]
\centering
\caption{Activity detection results on Charades (in \%). We report results using the same evaluation metric as in~\cite{sigurdsson2016asynchronous}.
}
\normalsize
\begin{tabular}{l || c c } 
\hline
 ~ & \multicolumn{2}{c}{mAP} \\ 
 ~ & \!\!\! standard \!\!\!\!\!\! & post-process \\ \hline
 
\!\!\!Random ~\cite{sigurdsson2016asynchronous} & 4.2 & 4.2 \\ 
\!\!\!RGB ~\cite{sigurdsson2016asynchronous} & 7.7 & 8.8 \\ 
\!\!\!Two-Stream ~\cite{sigurdsson2016asynchronous} & 7.7 & 10.0 \\ 
\!\!\!Two-Stream+LSTM ~\cite{sigurdsson2016asynchronous} \!\!\! & 8.3 &  8.8 \\ 
\!\!\!Sigurdsson et al.~\cite{sigurdsson2016asynchronous} & 9.6 & 12.1 \\ \hline
\!\!\!Single-stream \modelname~\cite{Xu2017} & 12.4 & 12.7\\
\!\!\!Two-stream R-C3D (Concat) & 12.4 & 12.6\\
\!\!\!Two-stream R-C3D (Sum) & 12.5 & 12.9\\
\!\!\!Single-stream R-C3D + OHEM & 13.0 & 13.3\\
\hline 
\end{tabular}
\label{res:charades}
\end{table}

\noindent{\textbf{Results}:}
The proposal evaluation results in terms of average AUC with 100 proposals per video for single-stream \modelname and two-stream \modelname models are shown in Table~\ref{res:proposal_charades}.
For this dataset also, both the two-stream extensions have almost the same performance as the single-stream R-C3D.

Table~\ref{res:charades} provides a comparative evaluation with various baseline models reported in~\cite{sigurdsson2016asynchronous}.
This approach~\cite{sigurdsson2016asynchronous} trains a CRF based video classification model (asynchronous temporal fields) and evaluates the prediction performance on 25 equidistant frames by making a multi-label prediction for each frame.
The activity localization result is reported in terms of mAP metric on these frames.
For a fair comparison, we map our activity segment prediction to 25 equidistant frames and evaluate using the same mAP evaluation metric.
A second evaluation strategy proposed in this work relies on a post-processing stage where the frame level predictions are averaged across 20 frames leading to more spatial consistency.
As shown in Table~\ref{res:charades}, our jointly optimized single stream model outperforms the asynchronous temporal fields model~\cite{sigurdsson2016asynchronous} as well as several baselines reported in the same paper~\cite{sigurdsson2016asynchronous}.
While the improvement over the standard method is as high as 2.8\%, the improvement after the post-processing is not as high.
One possible reason could be that our jointly optimized fully convolutional model captures the spatial consistency implicitly without requiring any manually-designed post-processing.

Similar to the proposal evaluation performance, the addition of the flow stream in the two-stream extensions does not affect the activity detection performance much on this dataset.
The relatively weak underlying RGB and flow C3D architectures and the poor illumination conditions on the dataset may be the reasons behind the stalled performances of the two-stream extensions.
Initialization with a better C3D classification model trained on indoor videos with these challenging conditions (\textit{e.g.}, the low illumination indoor scenes or the multi-label nature of the data) could possibly boost the performance.
Employing hard mining improves the activity detection result of single-stream \modelname by around 0.6\% in both standard and post-processing settings.
Figure~\ref{fig:vis_charades} shows some representative qualitative results of \modelname from one video in this dataset.
One of the major challenges of this dataset is the presence of a large number of temporally overlapping activities.
The results show that our model is capable of handling such scenarios to some extent.
This is achieved by the ability of the proposal subnet to produce possibly overlapping activity proposals and is further facilitated by the segment offset regression per activity class.

\subsection{Activity Detection Speed}
\label{exp:speed}

\begin{table}[!t]
\centering
\caption{Activity detection speed during inference. 
Note that, the two-stream models marked with asterisks * don't include the optical flow extraction time.
}
\normalsize
\begin{tabular}{l || c } 
\hline
~ & FPS   \\ \hline
\specialcellL{S-CNN~\cite{shou2016temporal}} & 60   \\ 
\specialcellL{DAP~\cite{escorcia2016daps}} & 134.1   \\ \hline
\specialcellL{Single-stream R-C3D (Titan X Maxwell)~\cite{Xu2017}} & 569  \\ \hline 
\specialcellL{Single-stream R-C3D (Titan X Pascal)~\cite{Xu2017}} & 1030  \\ 
\specialcellL{{Two-stream R-C3D (Concat) (Titan X Pascal)}$^*$} & 656  \\ 
\specialcellL{{Two-stream R-C3D (Sum) (Titan X Pascal)}$^*$} & 642  \\ 
\specialcellL{Single-stream R-C3D (OHEM) (Titan X Pascal)} & 1030  \\ \hline
\end{tabular}
\vspace{-0.2in}
\label{res:speed}
\end{table}

In this section, we compare detection speed of our model with two other state-of-the-art methods.
The comparison results are shown in Table~\ref{res:speed}.
S-CNN~\cite{shou2016temporal} uses a time-consuming sliding window strategy and predicts at 60 fps.
DAP~\cite{escorcia2016daps} incorporates a proposal prediction step on top of LSTM and predicts at 134.1 fps.
\modelname constructs the proposal and classification pipeline jointly and these two stages share the features making it significantly faster.
The speed of execution is 569 fps on a single Titan-X (Maxwell) GPU for the proposal and classification stages together.
On the upgraded Titan-X (Pascal) GPU, our inference speed reaches even higher (1030 fps).
One of the reasons of the speedup of \modelname over DAP may come from the fact that the LSTM recurrent architecture in DAP takes time to unroll, while \modelname directly accepts a wide range of frames as input and the convolutional features are shared by the proposal and classification subnets.
The activity detection speed drops to 656 fps for two-stream concat extension and 642 fps for two-stream sum extension on pre-computed optical flow images, because of the additional computation for the flow stream\footnote{Pre-computing the optical flow images takes extra time and runs at 32.4 fps on a single GPU using the code in~\cite{yuan2016temporal}.
The total activity detection speed for the two-stream models could be limited by the bottleneck of flow extraction. }.
OHEM extension of single-stream \modelname maintains almost the same activity detection speed as single-stream R-C3D, since inference in both the models are essentially the same.

\section{Conclusion}
\label{sec:conclusion}

In this paper, we explore a multi-stream network that augments RGB image features in \modelname with motion features for temporal action detection in untrimmed videos.
An online hard mining strategy improves the performance by intelligently training on hard examples.
Our proposed method not only detects activities more accurately, but also detects them fast.
Analysis of the experimental results and ablation studies indicates robustness of the method as well as significant result improvements over state-of-the-arts on three large-scale data sets with diverse characteristics.
The future directions include investigating the related applications of our \modelname framework in other computer vision tasks, \textit{e.g.} dense video captioning and localizing moments in videos.

\noindent{\textbf{Acknowledgement}:} 
This work is supported in part by the NSF and DARPA. 
We would thank Lu He for the meaningful discussions.

\ifCLASSOPTIONcaptionsoff
  \newpage
\fi

\bibliographystyle{IEEEtran}
\bibliography{egbib}


\begin{IEEEbiography}[{\includegraphics[width=1in,height=1.25in,clip]{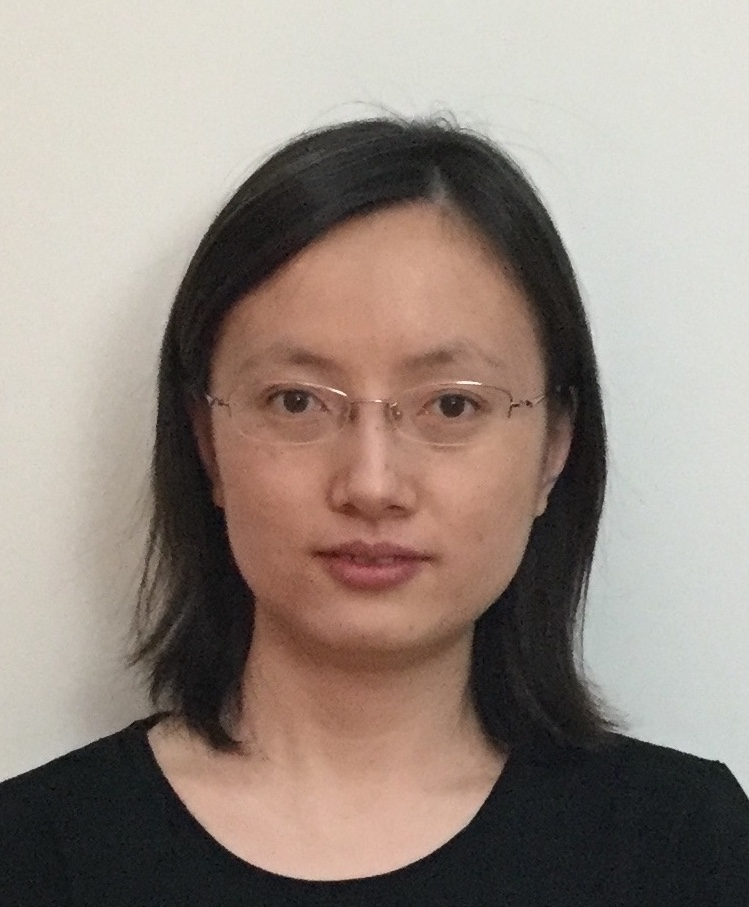}}]
{Huijuan Xu} is a postdoc at UC Berkeley. Huijuan received her PhD degree from computer science department at Boston University in 2018, advised by Professor Kate Saenko. Before that, she received her Bachelor's degree from Hefei University of Technology in 2009, and Master's degree from University of Chinese Academy of Sciences in 2012. Her research focuses on deep learning, computer vision and natural language processing, particularly in the area of visual question answering, video language description and activity detection. 
\end{IEEEbiography}

\begin{IEEEbiography}[{\includegraphics[width=1in,height=1.25in,clip]{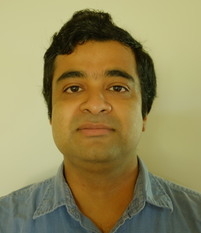}}]
{Abir Das} is an Assistant Professor of Computer Science and Engineering Department of IIT Kharagpur, India and the director of the Computer Vision and Intelligence Research (CVIR) group.
He received his B.E. in Electrical Engineering from Jadavpur University, India in 2007 and M.S. and Ph.D. in the same subject from University of California, Riverside in 2013 and 2015 respectively.
He was a postdoctoral researcher in the Computer Science department at Boston University.
His main research interests include computer vision, activity detection, person re-identification, explainable AI and bias in machine learning.
\end{IEEEbiography}

\begin{IEEEbiography}[{\includegraphics[width=1in,height=1in,clip]{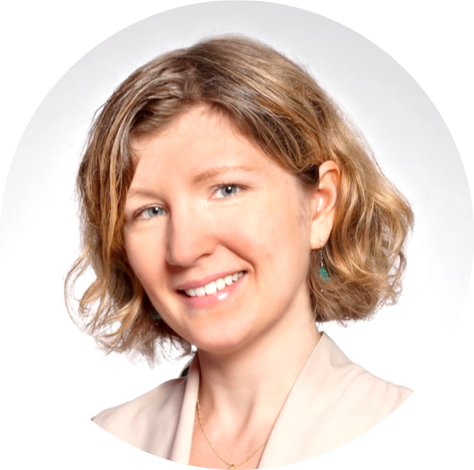}}] {Kate Saenko} is
an Associate Professor of Computer Science at Boston University, director of the Computer Vision and Learning Group and co-founder of the Artificial Intelligence Research (AIR) initiative. She was previously an Assistant Professor at the Computer Science Department at UMass Lowell, a Postdoctoral Researcher at the International Computer Science Institute (ICSI), Visiting Scholar at UC Berkeley in EECS and a Visiting Postdoctoral Fellow in the School of Engineering and Applied Science (SEAS) at Harvard University. Prof. Saenko's research interests are in the broad area of Artificial Intelligence with a focus on Adaptive Machine Learning, Learning for Vision and Language Understanding, and Deep Learning.
\end{IEEEbiography}

\end{document}